\documentclass{article}

\PassOptionsToPackage{numbers, compress}{natbib}
\usepackage[final]{neurips_2024}

\usepackage[position=bottom]{caption}
\usepackage{amsmath,amsfonts,bm}
\usepackage[utf8]{inputenc} % allow utf-8 input
\usepackage[T1]{fontenc}    % use 8-bit T1 fonts
\usepackage{url}            % simple URL typesetting
\usepackage{booktabs}       % professional-quality tables
\usepackage{nicefrac}       % compact symbols for 1/2, etc.
\usepackage{microtype}      % microtypography
\usepackage[numbers]{natbib}
\usepackage[dvipsnames,table,xcdraw]{xcolor}
\definecolor{cvprblue}{rgb}{0.21,0.49,0.74}
\usepackage{graphicx}
\usepackage{caption}
\usepackage{arydshln}
\usepackage{tikz}
\usepackage{wrapfig}
\usepackage{float}
\usepackage{multirow}
\usepackage{pifont}
\usepackage{makecell}
\usepackage{subfigure}
\usepackage{pdfpages}
\usepackage{xcolor}

\usepackage{tikz}
\usepackage{xcolor}

\usepackage{listings}

\usepackage{ulem}

\usepackage[pagebackref,breaklinks,colorlinks,citecolor=cvprblue]{hyperref}
\newcommand{\cmark}{\ding{51}}%
\newcommand{\xmark}{\ding{56}}
\usepackage{mdframed}
\definecolor{mygray}{gray}{0.97}
\colorlet{shadecolor}{mygray}
\newmdenv[%
  backgroundcolor=mygray, % Set the background color
  linewidth=0pt
]{newshaded}

\newcommand{\figref}[1]{Fig.~\ref{#1}}
\newcommand{\tabref}[1]{Tab.~\ref{#1}}

\newcommand{\secref}[1]{Sec.~\ref{#1}}

\title{Skywork UniPic 2.0: Building Kontext Model with Online RL for Unified Multimodal Model}

\author{
    \textnormal{Skywork Multimodality Team} \\
    \texttt{multimodal@skywork.ai}
    \\\\
    Project Page: \url{https://unipic-v2.github.io}
}

\begin{document}

\maketitle

\begin{abstract}

Recent advances in multimodal models have demonstrated impressive capabilities in unified image generation and editing. 
However, many prominent open-source models prioritize scaling model parameters over optimizing training strategies, limiting their efficiency and performance.
In this work, we present \textbf{UniPic2-SD3.5M-Kontext}, a 2B-parameter DiT model based on SD3.5-Medium, which achieves state-of-the-art image generation and editing while extending seamlessly into a unified multimodal framework. Our approach begins with architectural modifications to SD3.5-Medium and large-scale pre-training on high-quality data, enabling joint text-to-image generation and editing capabilities.
To enhance instruction following and editing consistency, we propose a novel Progressive Dual-Task Reinforcement strategy(PDTR), which effectively strengthens both tasks in a staged manner. We empirically validate that the reinforcement phases for different tasks are mutually beneficial and do not induce negative interference.
After pre-training and reinforcement strategies, UniPic2-SD3.5M-Kontext demonstrates stronger image generation and editing capabilities than models with significantly larger generation parameters—including BAGEL (7B) and Flux-Kontext (12B).
Furthermore, following the MetaQuery, we connect the UniPic2-SD3.5M-Kontext and Qwen2.5-VL-7B via a connector and perform joint training to launch a unified multimodal model \textbf{UniPic2-Metaquery}.
UniPic2-Metaquery integrates understanding, generation, and editing, achieving top-tier performance across diverse tasks with a simple and scalable training paradigm. This consistently validates the effectiveness and generalizability of our proposed training paradigm, which we formalize as \textbf{Skywork UniPic 2.0}. 

\end{abstract}

\section{Introduction}
\begin{figure}
    \centering
    \includegraphics[width=1\linewidth]{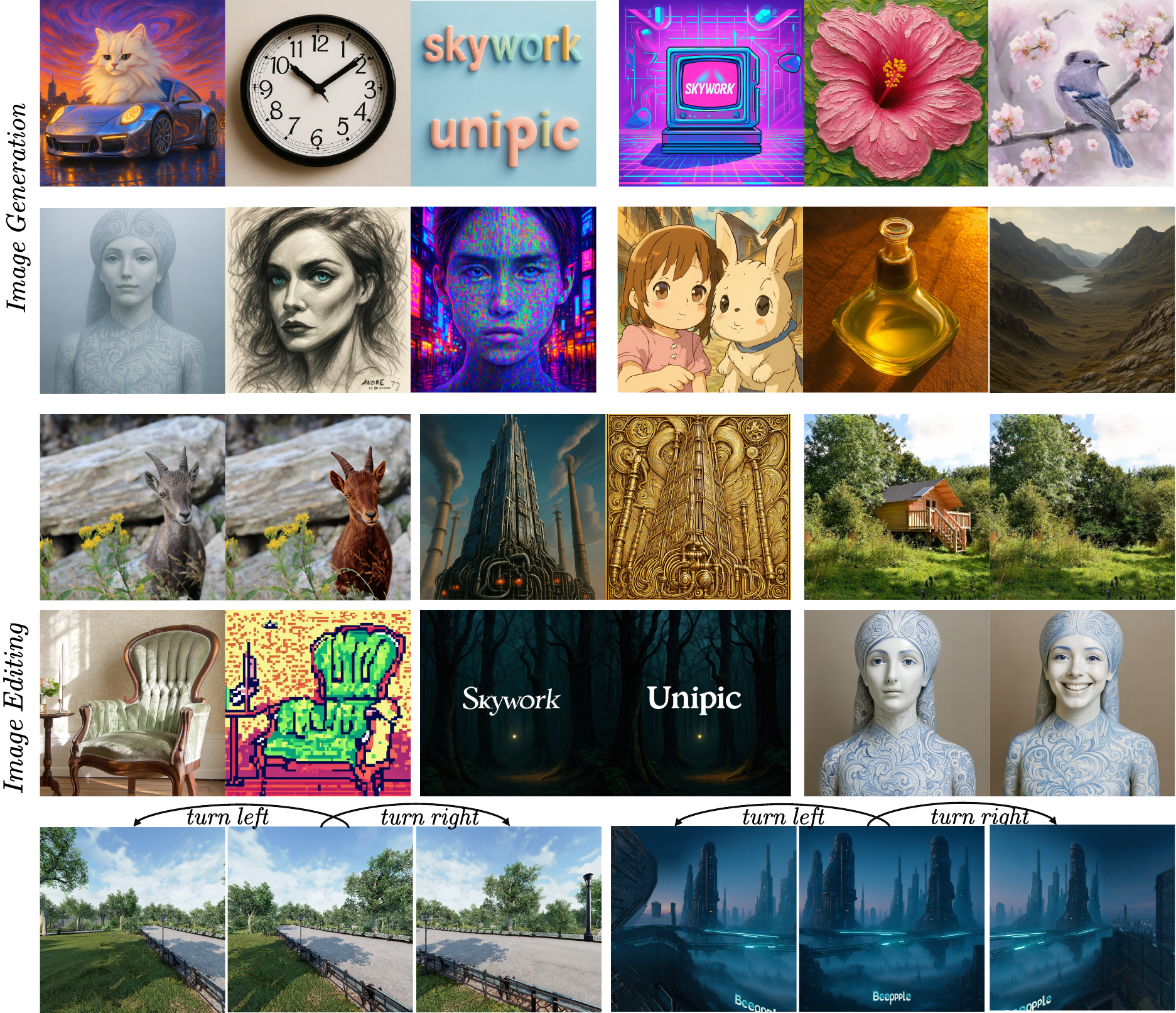}
    \caption{Showcase of the UniPic 2.0 in image generation and editing.}
    \label{fig:teaser}
    \vspace{-2em}
\end{figure}
Recent advances in multimodal generative models~\cite{batifol2025flux, deng2025bagel, wang2025ovis, wu2025omnigen2, lin2025uniworld, wu2025qwenimage} have demonstrated remarkable capabilities in unifying image generation and editing, offering both high visual fidelity and enhanced interactivity. Models such as BAGEL~\cite{deng2025bagel} and FLUX.1-Kontext-Dev~\cite{batifol2025flux} have pioneered unified architectures for text-to-image (T2I) synthesis and image editing tasks, advancing the development of intelligent multimodal systems. 
However, the image generation modules of these models typically contain tens of billions of parameters, leading to prohibitive computational costs and slow inference speeds.
Moreover, significant challenges persist in ensuring precise adherence to instructions in image generation and maintaining consistency in editing behaviors. An exclusive emphasis on parameter scaling, without corresponding advancements in training strategies, may prove suboptimal.

To evaluate these conjectures, we adopt the SD3.5-Medium~\cite{esser2024scalingrectifiedflowtransformers} architecture as our foundational model, which comprises a relatively compact 2B-parameter image generation module. We first introduce targeted architectural enhancements, followed by extensive pre-training on large-scale, high-fidelity datasets encompassing both image generation and editing tasks. This approach enables the resulting model to seamlessly support text-to-image (T2I) synthesis and image editing within a shared, integrated framework.
Furthermore, by employing progressive resolution training in conjunction with balanced data sampling across a wide spectrum of aspect ratios and resolutions, the proposed model attains native support for dynamic resolution generation during both training and inference, which is pivotal for practical use in real-world scenarios.

Although reinforcement learning~\cite{rafailov2023direct,wallace2024diffusion,dong2023raft,wallace2024diffusion,black2023training,fan2023reinforcement} has been increasingly adopted to align text-to-image generation with human preferences, its application to jointly optimize generation and editing in a unified model remains unexplored. 
Due to significant differences in input modalities, output distributions, and evaluation criteria between the two tasks, naively optimizing them jointly often leads to gradient conflicts or performance degradation—a classic multitask learning dilemma where ``optimizing one capability compromises the other.''
To overcome this challenge and enhance both instruction following in image generation and editing consistency, we follow Flow-GRPO~\cite{liu2025flow} to propose Progressive Dual-Task Reinforcement (PDTR), a novel post-training paradigm that leverages Group Relative Policy Optimization (GRPO)~\cite{shao2024deepseekmath} for online reinforcement learning.
Specifically, PDTR first reinforces the image editing task independently, followed by a second phase focusing on T2I generation. 
Our experiments demonstrate that the two tasks can be iteratively improved in a synergistic manner without negative interference, effectively resolving the conflict in multi-task RL. 
In the image editing enhancement phase, we leveraged both the self-trained Skywork-EditReward model and the online GPT-4.1~\cite{gpt4-1} system as reward evaluators. For the image generation enhancement phase, we employed GenEval~\cite{ghosh2023geneval} as a verifiable reward signal, complemented by classical detector~\cite{cheng2022masked,chen2019mmdetection} to automatically assess compositional accuracy and instruction adherence. This multi-faceted evaluation strategy reinforces the model’s robustness and reliability in handling complex and fine-grained prompts.

Building upon large-scale pre-training and PDTR optimization on a modified SD3.5M architecture, we introduce UniPic2-SD3.5M-Kontext, a unified model that demonstrates strong capabilities in both image generation and editing. Comprehensive evaluations indicate that UniPic2-SD3.5M-Kontext attains leading performance across multiple benchmarks: it outperforms most existing unified frameworks on GenEval in instruction-following accuracy and establishes new records in editing quality, thereby substantiating its superior generation fidelity, instruction comprehension, and editing consistency.

Motivated by the recent advances in MetaQuery~\cite{pan2025transfermodalitiesmetaqueries}, we investigate the integration of UniPic2-SD3.5M-Kontext, which has strong image generation and editing capabilities, with Qwen2.5-VL-7B~\cite{bai2025qwen2}, which excels at multimodal understanding, to construct a unified framework for multimodal understanding, generation, and editing. In the first stage, we freeze the parameters of both the SD3.5-Medium and Qwen2.5-VL models, and pre-train a 24-layer connector module using large-scale text-to-image datasets. In the second stage, we substitute SD3.5-Medium with UniPic2-SD3.5M-Kontext, unfreeze the connector parameters, and jointly fine-tune the connector and Kontext components on high-quality image generation and editing datasets. This yields UniPic2-MetaQuery, a unified multimodal model capable of robust understanding, high-fidelity image generation, and consistent image editing. The proposed training paradigm is simple yet highly scalable, and achieves state-of-the-art results across diverse benchmarks, highlighting its strong modularity and extensibility.

We integrate UniPic2-SD3.5M-Kontext and UniPic2-MetaQuery into Skywork UniPic 2.0, an efficient generative framework for unified multimodal modeling that is designed to enhance speed, efficiency, and generalization, with its capabilities illustrated in Fig.~\ref{fig:teaser}. All models and source code are publicly released to foster reproducibility and accelerate progress in efficient multimodal generation. Our results demonstrate that, through deliberate architectural design, targeted pretraining strategies, and coordinated reinforcement learning, lightweight models can outperform substantially larger counterparts in generation fidelity, instruction compliance, and inference efficiency. Skywork UniPic 2.0 thus offers a practical and scalable pathway toward deployable, high-performance multimodal intelligence.

Key contributions of this work are summarized as follows:
\begin{itemize}
    \item We present \textbf{UniPic2-SD3.5M-Kontext}, a lightweight unified model for image generation and editing, enabled by large-scale pretraining, achieving leading performance under high inference speed.
    
    \item We introduce \textbf{Progressive Dual-Task Reinforcement (PDTR)}, the first strategy to enable synergistic improvement of image generation and image editing through staged RL, without cross-task interference—significantly boosting instruction following of generation and editing consistency.
    
    \item We introduce \textbf{UniPic2-Metaquery}, a general and modular paradigm for unified multimodal modeling, which enables end-to-end integration of understanding, generation, and editing through a parameter-efficient connector-based training strategy, achieving SOTA performance and strong generalization across tasks.
\end{itemize}

\section{Related Work}
\subsection{Image Generation}
Diffusion models~\cite{ho2020denoising, rombach2022highresolutionimagesynthesislatent, ramesh2021zero, 2023SDXL, 2024pixartsigma, ramesh2022hierarchical, dalle3, peebles2023scalable, 2024hunyuandit} have emerged as the dominant paradigm for image generation, achieving high-quality visual synthesis conditioned on text prompts. 
These models are typically instantiated in the latent space of variational autoencoders~\cite{kingma2013auto, van2017neural} (VAE), significantly reducing computational and memory overhead.
Recent advances~\cite{esser2024scalingrectifiedflowtransformers, xie2024sanaefficienthighresolutionimage, flux2024, ma2024sit,xue2025dancegrpounleashinggrpovisual} adopt Flow Matching (FM), which replaces explicit diffusion simulation with an ODE-driven continuous transformation from noise to data, effectively reducing the diffusion processes and permitting more direct probability transport paths to improve efficiency.
Meanwhile, diffusion transformers~\cite{vaswani2017attention} (DiTs) have been adopted as the architectural backbone in state-of-the-art frameworks~\cite{peebles2023scalable, flux2024, ma2024sit, zhuo2024lumina, xie2025sana}, surpassing U-Net backbone~\cite{ronneberger2015u} in superior scalability and generation quality.

\subsection{Image Understanding}
Multimodal large language models~\cite{zhu2023minigpt4enhancingvisionlanguageunderstanding, liu2023visualinstructiontuning, liu2024improved, li2024llava, zhu2025internvl3exploringadvancedtraining, wang2024qwen2, li2024mini, wu2024deepseek, pure} (MLLMs) combine visual encoders~\cite{clip,zhai2023sigmoid, tschannen2025siglip} with language models~\cite{yang2024qwen2, touvron2023llama, touvron2023llama2} to enable a wide range of image understanding tasks, including visual reasoning, dialogue, and instruction following. The visual encoders are typically pretrained through contrastive vision-language alignment, allowing the model to capture semantic concepts.
While MLLMs demonstrated strong general and fine-grained image understanding capabilities by incorporating visual signals into LLMs, their output is primarily text, lacking the ability to produce visual content.

\subsection{Unified Models}
Efforts to unify multimodal understanding and generation generally follow two paradigms.
The first builds native multimodals from scratch~\cite{team2024chameleon,wu2024vila,xie2024show,wu2024janus,chen2025janusprounifiedmultimodalunderstanding,li2024synergen,deng2025bagel}, and struggles to accommodate both tasks due to the inherently different granularity. To address this, some adopt disentangled visual encoders~\cite{wu2024janus, chen2025janusprounifiedmultimodalunderstanding} or mixture-of-experts architectures~\cite{li2024synergen,deng2025bagel,shukor2025scaling}.
The more resource-efficient line of work connects pretrained LLMs and diffusion generators using LLM hidden-states~\cite{shukor2025scaling, wang2024illume,tong2024metamorph,wu2025harmonizing, lin2025uniworld, wu2025omnigen2} or learnable queries~\cite{chen2025blip3ofamilyfullyopen, pan2025transfermodalitiesmetaqueries, wu2025openuni}. Among these works, MetaQuery~\cite{pan2025transfermodalitiesmetaqueries}and BLIP3-o~\cite{chen2025blip3ofamilyfullyopen} build unified frameworks upon frozen multimodal LLMs, effectively transferring the knowledge learned in understanding tasks to visual generation. UniPic 2.0 adopts the query-based design considering its training efficiency and excellent performance.
\section{Method}
\begin{figure}
    \centering
    \includegraphics[width=1\linewidth]{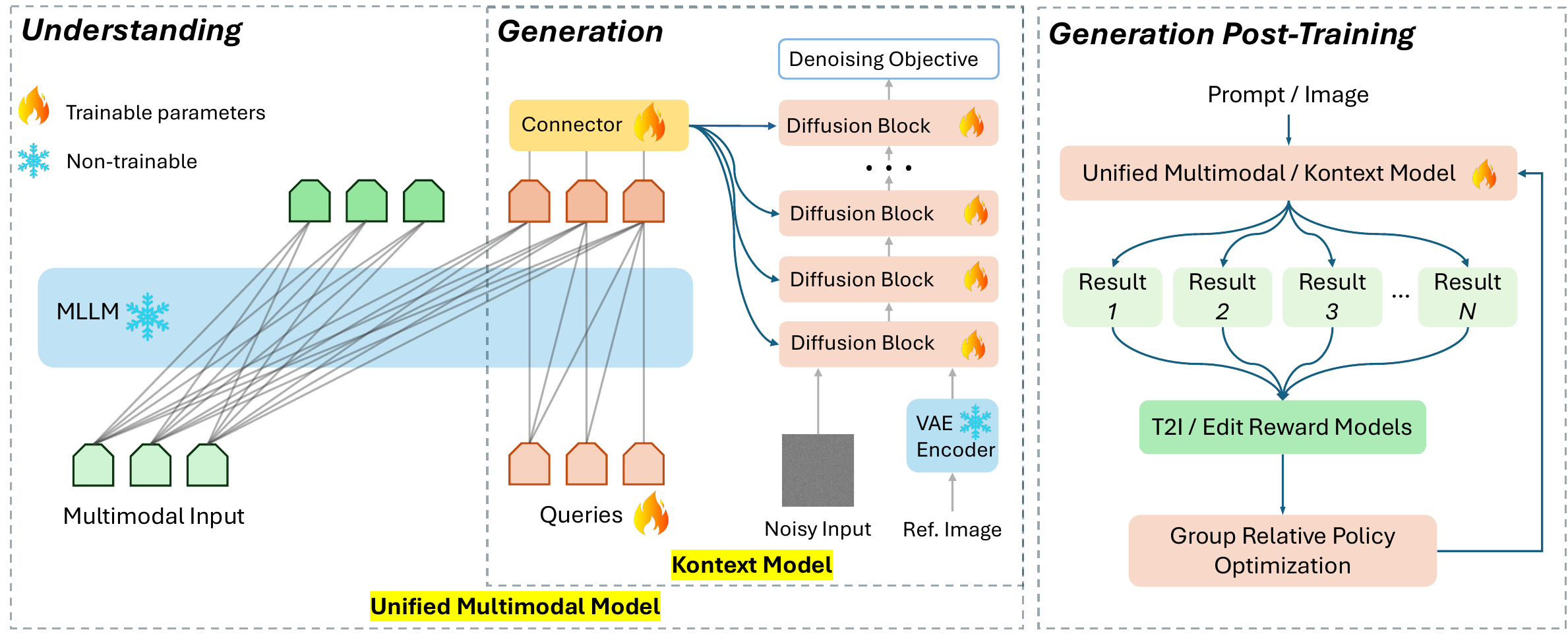}
    \caption{The overall pipeline of UniPic 2.0.}
    % \caption{Overview of UniPic 2.0. For image generation, the learnable queries are used to gather conditional information in the forward pass through the MLLM, which are then processed by the connector before being fed to the DiT module. For image understanding, we fully preserve the MLLM's original ability thanks to the `frozen' design choice.}
    \label{fig:pipeline}
    \vspace{-1em}
\end{figure}
Different from UniPic 1.0~\cite{wang2025skyworkunipicunifiedautoregressive}, which was trained from scratch as an autoregressive unified model, UniPic 2.0 adopts a MetaQuery-style design paradigm. This approach enables us to combine a frozen multimodal large language model with a pretrained diffusion generator, achieving efficient training while retaining strong performance in both understanding and generation tasks.
As shown in Fig.~\ref{fig:pipeline},  UniPic 2.0 follows the resource-efficient design of MetaQuery ~\cite{pan2025transfermodalitiesmetaqueries} to build UniPic 2.0. The overall architecture mainly comprises an off-the-shelf mutlimodal large language model (MLLM) and a diffusion transformer (DiT), bridged by a set of learnable queries and a connector. To preserve the MLLM's ability in multimodal understanding, we freeze its parameters throughout the training process. We first pre-train UniPic 2.0 Kontext model on text-to-image and image editing datasets, and align the Kontext model and the MLLM model to achieve a unified multimodal model (Sec.~\ref{sec:pre_train}). Then the models are further optimized via GRPO in a post-training process (Sec.~\ref{sec:post_train}).
In our implementation, we choose Qwen2.5VL-7B~\cite{bai2025qwen2} and SD3.5-Medium~\cite{esser2024scalingrectifiedflowtransformers} as the instantiations of the MLLM and DiT. 

\subsection{Pre-training}\label{sec:pre_train}
In the pre-training stage, we extend the text-to-image DiT to support image editing, optimize the connector modules to align the MLLM with the DiT, and jointly fine-tune both the connector and DiT to enable unified multimodal modeling.

\paragraph{Kontext Model.} Since the original DiT backbone in SD3.5-Medium is trained solely for text-to-image generation, we retrain it on a mixed corpus of image generation and image editing datasets. Prior works~\cite{lin2025uniworld, deng2025bagel, wu2025omnigen2} have highlighted the role of VAE latents in preserving structural and textural fidelity, so we inject the reference images’ VAE latents into DiT’s self-attention layers to enhance editing capability.
In this design, the model conditions simultaneously on textual instructions and reference images. The text encoder transforms the instruction into an embedding, while the VAE encodes the reference image into a latent representation, which is then projected into context tokens. These tokens are concatenated with the target image’s noise tokens to form a single input sequence, where the model’s positional encoding distinguishes between reference-image tokens and target-image tokens.
We refer to this retrofitted DiT (SD3.5-Medium) with joint generation–editing capability as UniPic2-SD3.5M-Kontext. During training, image generation and editing batches alternate to jointly optimize both tasks. Resolutions are sampled to include common aspect ratios (1:1, 4:3, 3:2, 16:9) to avoid scale or ratio bias, thus improving generalization to varied compositions and scenarios.

\paragraph{Align MLLM to Kontext Model.}
Following the MetaQuery~\cite{pan2025transfermodalitiesmetaqueries} paradigm, we align the pre-trained Kontext model with Qwen2.5-VL via a transformer-based connector. As emphasized in MetaQuery, a high-capacity connector is critical for effectively bridging the MLLM and DiT; accordingly, our design employs a 24-layer transformer with approximately 1 billion parameters.
During alignment, we freeze both the MLLM and DiT, and train the connector and learnable queries on large-scale text-to-image data.
Empirically, we observe that the DiT module prompted by our learnable queries and connector exhibits even better text-to-image generation performance compared to using its original text encoders, e.g., T5~\cite{raffel2020exploring} in SD3.5.

\paragraph{Unified Multimodal Model.} With the connecting modules and the Kontext model obtained in the previous stages, we jointly fine-tune the learnable queries, connector and DiT while freezing the MLLM for unified image understanding, generation and editing. We refer to our unified model as \emph{UniPic2-MetaQuery}.
For image editing tasks, it is noteworthy that we harness the MLLM's image understanding ability to provide semantic-rich conditioning by feeding both text instructions and reference images to the MLLM, in addition to the VAE latents that are passed to the DiT's self-attention modules.

\subsection{Post-training}\label{sec:post_train}
% \subsubsection{Group Relative Policy Optimization (GRPO)}
\paragraph{Group Relative Policy Optimization (GRPO).}
Given text hidden state $h$, the flow model samples a group of $G$ images $\left\{x_{0}^{i}\right\}_{i=1}^{G}$ and the corresponding trajectories $\left\{x_{T}^{i}, x_{T-1}^{i}, \ldots, x_{0}^{i}\right\}_{i=1}^{G}$, following the Flow-GRPO~\cite{liu2025flow}. Within each group, the advantage of the $i$-th image can be formulated as:
\begin{equation}
    A_{i}=\frac{R\left(x_{0}^{i}, h\right)-\operatorname{mean}\left(\left\{R\left(x_{0}^{i}, h\right)\right\}_{i=1}^{G}\right)}{\operatorname{std}\left(\left\{R\left(x_{0}^{i}, h\right)\right\}_{i=1}^{G}\right)},
\end{equation}
where $R$ denotes the reward model and GRPO employs a clipped objective. The GRPO training objective with a KL penalty term is given:
\begin{align}
\mathcal{L}_{\mathrm{GRPO}}(\theta)= & \mathbb{E}_{h \sim \mathcal{D},\left\{x_{T}^{i}, \ldots, x_{0}^{i}\right\}_{i=1}^{G} \sim \pi_{\theta}} \nonumber \\
& \frac{1}{G} \sum_{i=1}^{G} \frac{1}{T} \sum_{t=0}^{T-1}\left(\min \left(r_{t}^{i}(\theta) A_{i}, \operatorname{clip}\left(r_{t}^{i}(\theta), 1-\epsilon, 1+\epsilon\right) A_{i}\right)-\beta D_{K L}\left(\pi_{\theta} \| \pi_{\mathrm{ref}}\right)\right),
\end{align}
where $r_{t}^{i}(\theta)=\frac{p_{\theta}\left(x_{t-1}^{i} \mid x_{t}^{i}, h\right)}{p_{\theta_{\text {old }}}\left(x_{t-1}^{i} \mid x_{t}^{i}, h\right)}$.

% \subsubsection{}
\paragraph{Progressive Dual-Task Reinforcement (PDTR).}
Currently, no effective paradigm exists for applying reinforcement learning (RL) to jointly optimize text-to-image (T2I) generation and image editing within a unified model.
To address this challenge, we propose the Progressive Dual-Task Reinforcement (PDTR), the first strategy to enable synergistic reinforcement learning of T2I and editing in a shared diffusion model. The core idea of PDTR is to decouple the optimization order of tasks through a staged, progressive training schedule, allowing the model to incrementally improve performance on a new task while preserving its existing capabilities, thereby avoiding cross-task interference. PDTR consists of two sequential stages as follows.

\paragraph{Stage 1: Image Editing Reinforcement.}
We first conduct independent reinforcement learning for image editing. Given an input image and a textual edit instruction, the goal is to generate outputs that are both semantically consistent and visually natural. Following Flow-GRPO, we adopt GRPO as the RL algorithm and design multi reward signals. We systematically validate the effectiveness of two editing reward signals: our self-trained Skywork-EditReward model and online GPT-4.1 evaluation. After this stage, the model progressively masters precise execution of complex edits while preserving source image structures.

\paragraph{Stage 2: Text-to-Image Reinforcement.}
Building upon the editing-enhanced model from Stage 1, we proceed to reinforce the T2I generation capability. Using GRPO with verifiable rewards, the model learns to generate images with richer semantic structures. Notably, we introduce compositionality metrics from the GenEval benchmark (e.g., attribute-object binding, spatial relation reasoning) as reward signals, combined with automated detection methods (e.g., object detection and layout analysis) for scalable online feedback. This stage significantly improves instruction adherence and semantic precision. 

Critically, we conduct systematic evaluations across multiple training rounds to assess the interaction between the two stages. Experimental results show that reinforcing one task improves performance on the other: for instance, editing-focused RL enhances T2I quality, and vice versa. Moreover, reinforcing T2I after editing does not degrade editing performance. This positive cross-task transfer validates the effectiveness of PDTR: with proper task scheduling and optimization design, generation and editing can undergo non-adversarial co-evolution within a shared model architecture.

\paragraph{Skywork-EditReward vs. Online GPT-4.1.}
To enable more precise feedback for editing reinforcement learning, we develop Skywork-EditReward, a specialized reward model tailored for image editing quality assessment. 
% Our approach consists of two key components:
% \textit{GPT-4.1-generated training samples}: 
Firstly, we leverage our pre-trained UniPic2-SD3.5M-Kontext model to generate 333k editing results. Through carefully designed stable and effective evaluation templates, we employ GPT-4.1 to provide multi-dimensional scoring aligned with human aesthetic standards, evaluating aspects such as instruction following accuracy and image quality.
% \textit{Multimodal architecture training}: 
Then, based on Qwen2.5-VL-7B~\cite{wang2024qwen2}, we train Skywork-EditReward using supervised learning with regression loss, enabling the model to predict quality scores that are highly consistent with GPT-4.1 evaluations.
In reinforcement learning, Skywork-EditReward captures subtle quality differences that traditional metrics cannot quantify, significantly improving the naturalness and consistency of editing results. This provides an effective technical pathway for future reinforcement learning post-training in image editing tasks.

\begin{figure}[!htbp]
    \centering
    % \vspace{-2em}
    \makebox[\textwidth]{\includegraphics[width=1\linewidth]{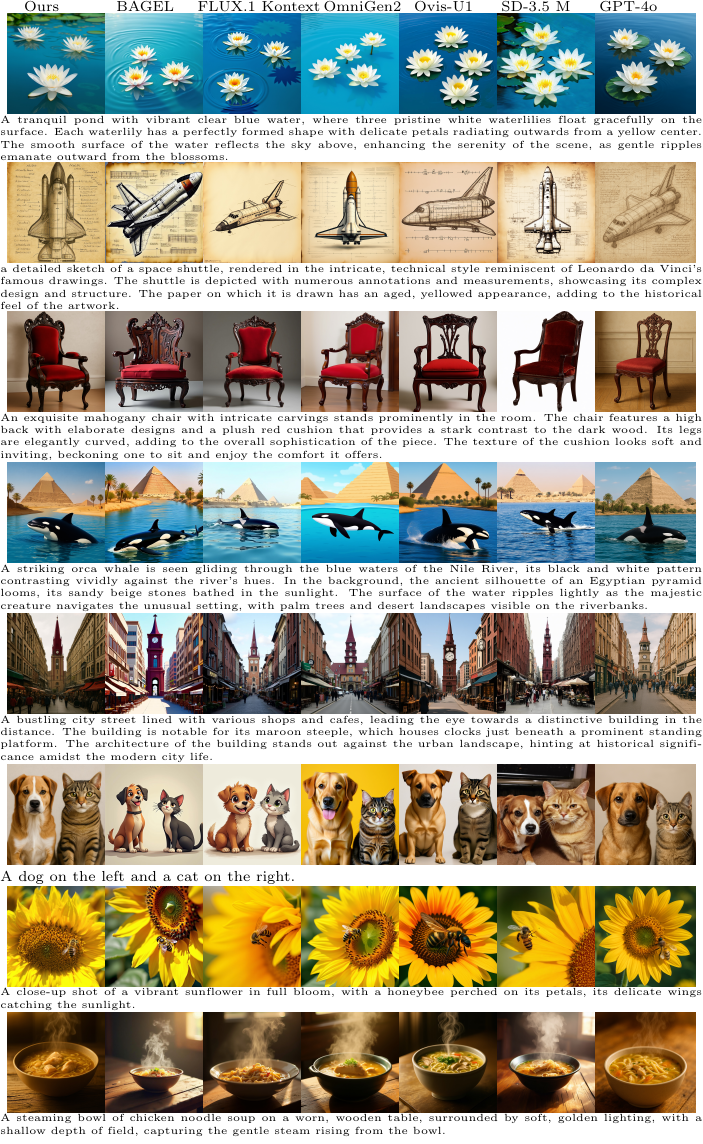}}
    \caption{Qualitative comparison of text-to-image generation results.}
    \label{fig:vis_gen}
\end{figure}

\section{Experiments}

\subsection{Setup}

\paragraph{Pre-training.} For aligning the MLLM and the DiT, we train the connecting modules on 150M image-text pairs for $500K$ steps with a global batch size of $1,024$ and a learning rate of $1e-4$. The training data includes 30M data samples from open-source datasets (CC12M~\cite{changpinyo2021conceptual}, Megalith10M~\cite{madebyollin_megalith_10m}, RedCaps~\cite{desai2021redcaps} and Laion-Aesthetics~\cite{dclure_laion_aesthetics_12m_umap}) and 120M internal synthetic images. For UniPic2-SD3.5M-Kontext, we train the model on 5M image editing data samples and 6M text-to-image data samples for $200K$ steps with a global batch size of 128 and a learning rate of $4\times 10^{-5}$. The image editing data includes editing pairs collected from UniWorld-V1~\cite{lin2025uniworld}, OmniGen2~\cite{wu2025omnigen2}, NHR-Edit~\cite{kuprashevich2025nohumansrequired}, ShareGPT4o-Image~\cite{chen2025sharegpt} and GPT-Image-Edit-1.5M~\cite{wang2025gpt} as well as 450K internal data samples. The text-to-image data mainly comprises 5.6M internal high-quality real images and 400K public data from BLIP3o~\cite{chen2025blip3ofamilyfullyopen} and ShareGPT4o-Image~\cite{chen2025sharegpt}. It is noteworthy that the same data and hyperparameters are used for training our unified model in the last stage.
For all stages in pre-training, we use cosine learning rate annealing and AdamW optimizer with $\beta_1 = 0.9$, $\beta_2 = 0.95$, $\epsilon = 1 \times 10^{-8}$ and a weight decay of $0.05$.

\paragraph{Post-Training.}
For post-training, all hyperparameters are kept fixed across different reinforcement learning stages. We use a sampling timestep of $T = 10$, a group size $G = 16$, noise level $a = 0.7$, and image resolution of $512 \times 512$. The KL ratio $\beta$ is set to $0.04$. We employ LoRA for parameter-efficient fine-tuning, with rank $r = 32$ and scaling factor $\alpha = 64$. Training uses a learning rate of $3 \times 10^{-4}$ and a global batch size of 256. We use the Adam optimizer with $\beta_1 = 0.9$, $\beta_2 = 0.999$, and $\epsilon = 1 \times 10^{-8}$. For the editing reinforcement data, we used the ShareGPT4o-Image~\cite{chen2025sharegpt}. As for the text-to-image enhancement phase, we employed the same Geneval training set as Flow-GRPO~\cite{liu2025flow}.

\subsection{Main results.} \tabref{tab:main_res} summarizes our main comparative results, evaluating our models against other advanced methods on image generation, image editing, and multimodal understanding. We compare with GPT-4o~\cite{hurst2024gpt}, Emu3~\cite{wang2024emu3}, Janus-Pro~\cite{chen2025janusprounifiedmultimodalunderstanding}, Blip3-o~\cite{chen2025blip3ofamilyfullyopen}, BAGEL~\cite{deng2025bagel}, UniWorld-V1~\cite{lin2025uniworld}, OmniGen2~\cite{wu2025omnigen2}, and Ovis-U1~\cite{wang2025ovis}. The results clearly demonstrate the exceptional performance of our approach across major benchmarks, highlighting its powerful unified capabilities. Notably, we achieve a leading performance with fewer parameters. Detailed comparisons for text-to-image generation, image editing, and multimodal understanding are provided in subsequent subsections. We further present ablation studies in \secref{subsec:aba} and analyze failure cases in \secref{subsec:fail}.

\begin{table*}[!htbp]
    \centering
    \setlength{\tabcolsep}{0.7pt}
    \renewcommand{\arraystretch}{1.1}
    % \scriptsize
    \footnotesize
    % \small
    \caption{\textbf{Comparisons on image understanding, generation, and editing.} `Generation' and `Editing' refer to models specialized in image generation and image editing, respectively, while `Unified' denotes a model that has both understanding and generation capabilities. `×' indicates the model is incapable of performing the task. `$^{\dagger}$' indicates results obtained using the official OpenAI API. `*' denotes generation results for SD3.5-Medium are reported based on our evaluation.
    }
    \begin{tabular}{cccccccccc}
        \toprule
        \multirow{2}{*}{\textbf{Type}} & \multirow{2}{*}{\textbf{Model}} & \multirow{2}{*}{\textbf{$\#$ Params.}} &\multicolumn{2}{c}{\textbf{Generation}} & \multicolumn{2}{c}{\textbf{Editing}} & \multicolumn{3}{c}{\textbf{Understanding}} \\ 
        & & & GenEval & DPG & GEdit-En & Imgedit & MMBench & MMMU & MM-Vet \\ 
        \midrule
        \multirow{5}{*}{\rotatebox{90}{\textit{Generation}}}
        % & Emu$3$-Gen~\cite{wang2024emu3} & - & 0.54 & 80.60 & × & × & × & × & ×\\
        & SDXL~\cite{2023SDXL}& - & 0.55 & 74.65 & × & × & × & × & × \\
        & DALL-E $3$~\cite{dalle3} & - & 0.67 & 83.50 & × & × & × & × & ×\\
        & FLUX.1-dev~\cite{flux2024} & - & 0.67 & 84.00 & × & × & × & × & × \\
        & FLUX.1 Kontext~\cite{batifol2025flux}& - & - & - & 6.26 & 3.52 & × & × & ×\\
        & SD3.5-Medium*~\cite{esser2024scalingrectifiedflowtransformers} & - & 0.65 & 83.86 & × & × & × & × & × \\
        % & SD3-Medium~\cite{esser2024scalingrectifiedflowtransformers} & - & 0.74 & 84.08 & × & × & × & × & ×\\
        \midrule
        \multirow{4}{*}{\rotatebox{90}{\textit{Editing}}}
        % \multicolumn{9}{c}{\textbf{\textit{Image Editing}}} \\
        & AnyEdit~\cite{yu2025anyedit} & - & × & ×& 3.21 & 2.45 & × & × & × \\
        & Instructuct-P2P~\cite{brooks2023instructpix2pix} & - & × & × & 3.68 & 1.88 & × & × & × \\
        & MagicBrush~\cite{zhang2023magicbrush} & - & × & × & 4.52 & 1.90 & × & × & × \\
        & Step1X-Edit~\cite{liu2025step1x} & - & × & × & 6.97 & 3.06 & × & × & × \\
        \midrule
        \multirow{9}{*}{\rotatebox{90}{\textit{Unified}}}       
        & Emu$3$ \cite{wang2024emu3} & 8B & 0.66 & 80.60 & - & - & 58.5 & 31.6 & 37.2\\
        & Janus-Pro~\cite{chen2025janusprounifiedmultimodalunderstanding} & 7B & 0.80 & 84.19 & × & × & 75.5 & 36.3 & 39.8\\
        & MetaQuery-XL~\cite{pan2025transfermodalitiesmetaqueries} & 7B + 1.6B & 0.80 & 82.05 & - & - & 83.5 & 58.6 & 66.6 \\
        & UniWorld-V1~\cite{lin2025uniworld} & 7B + 12B & 0.84 & 81.38 & 4.85 & 3.26 & 83.5 & 58.6 & 67.1 \\
        & Blip3-o-8B~\cite{chen2025blip3ofamilyfullyopen} & 7B + 1.4B & 0.84 & 81.60 & × & × & 83.5 & 58.6 & 66.6 \\
        & OmniGen2~\cite{wu2025omnigen2}& 3B + 4B & 0.86 & 83.57 & 6.42 & 3.44 & 79.1& 53.1 & 61.8 \\
        & BAGEL~\cite{deng2025bagel} & 7B + 7B & 0.88 & 85.07 & 6.52 & 3.20 & 85.0 & 55.3 & 67.2 \\
        & Ovis-U1~\cite{wang2025ovis} & 2.4B + 1.2B & 0.89 & 83.72 & 6.42 & 4.00 & 77.8 & 51.1 & 66.7 \\
        & GPT-4o~\cite{hurst2024gpt} & - & 0.84 & 85.15 & 7.53 & 4.20 & 86.0 & 72.9 & 76.9\\
        \midrule
        \multirow{4}{*}{\rotatebox{90}{\textit{Ours}}}
        & UniPic2-SD3.5M-Kontext & 2B & 0.89 & 84.23 & 6.59 & 4.00 & × & × & × \\
        & UniPic2-SD3.5M-Kontext $^{\dagger}$ & 2B & 0.89 & 84.23 & 6.74 & 4.02 & × & × & × \\
        & UniPic2-Metaquery & 7B + 2B & 0.90 & 83.79 & 6.87 & 4.03 & 83.5 & 58.6 & 67.1 \\
        & UniPic2-Metaquery $^{\dagger}$ & 7B + 2B & 0.90 & 83.79 & 7.10 & 4.06 & 83.5 & 58.6 & 67.1 \\
    \bottomrule
    \end{tabular}
    \label{tab:main_res}
\end{table*}

\paragraph{Text-to-Image Generation.}\label{subsec:t2i}
To evaluate the text-to-image generation capabilities of our method, we employ two established benchmarks: GenEval~\cite{ghosh2023geneval}, which assesses the ability to generate images with accurate object attributes, counts, positions, and colors; and DPG-Bench~\cite{hu2024ella}, which evaluates fine-grained semantic alignment using long and dense prompts.

As shown in \tabref{tab:main_res}, our lightweight model UniPic2-SD3.5M-Kontext (0.89) outperforms both specialized image generation models like FLUX.1-dev (0.67) and SD3.5-Medium (0.65), and larger unified models including BAGEL (0.88), Blip3-o-8B (0.84), and MetaQuery-XL (0.80). UniPic2-Metaquery archives an overall score of 0.90. On the DPG-Bench benchmark, which contains lengthy and free-form text prompts, UniPic2-SD3.5M-Kontext (84.23) also surpasses specialized generation methods and achieves performance comparable to much larger models, such as BAGEL with 7 billion generation parameters. We further conduct a qualitative comparison between BAGEL, FLUX.1 Kontext, OmniGen2, Ovis-U1, SD3.5-Medium and GPT-4o. As illustrated in \figref{fig:vis_gen}, our method delivers faithful results that capture the specified counts, positions, colors, and styles described in the prompts.

\begin{figure}
    \centering
    \includegraphics[width=1\linewidth]{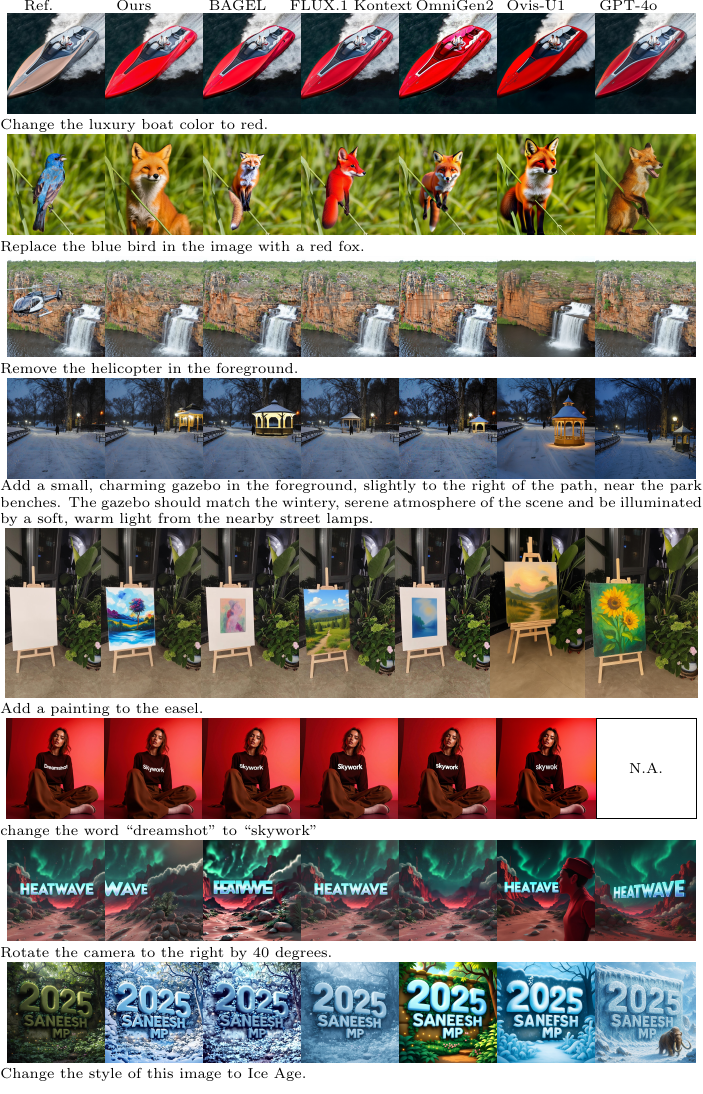}
    \caption{Qualitative comparison of image editing results.}
    \label{fig:vis_edit}
\end{figure}

\paragraph{Image Editing.}\label{subsec:edit}
Our UniPic2-SD3.5M-Kontext, with only 2B parameters in the generation module, achieve scores of (6.59, 4.00) on the GEdit and ImgEdit editing benchmarks, which outperforms the 12B Flux-Kontext (6.26, 3.52) and the 7B BAGEL (6.52, 3.20).
Furthermore, as shown in \tabref{tab:main_res}, benefiting from the internal knowledge and reasoning abilities of Qwen2.5-VL~\cite{bai2025qwen2}, UniPic2-Metaquery achieves a comparable score (6.87) to Step1X-Edit (6.97) on the GEdit-En benchmark, and even surpasses a range of specialized models, including FLUX.1 Kontext (3.52) and Step1X-Edit (3.06), as well as unified models such as Ovis-U1 (4.00) and OmniGen2 (3.44) on ImgEdit, achieving 4.03. As illustrated in \figref{fig:vis_edit}, our model produces more consistent results in removing, replacing, or adding objects and applying style changes, while preserving the rest of the reference image. `N.A.' indicates GPT-4o declined to produce results as the specified image editing involved brand or logo modifications may violate its content policies.

\begin{figure}
    \centering
    \includegraphics[width=1\linewidth]{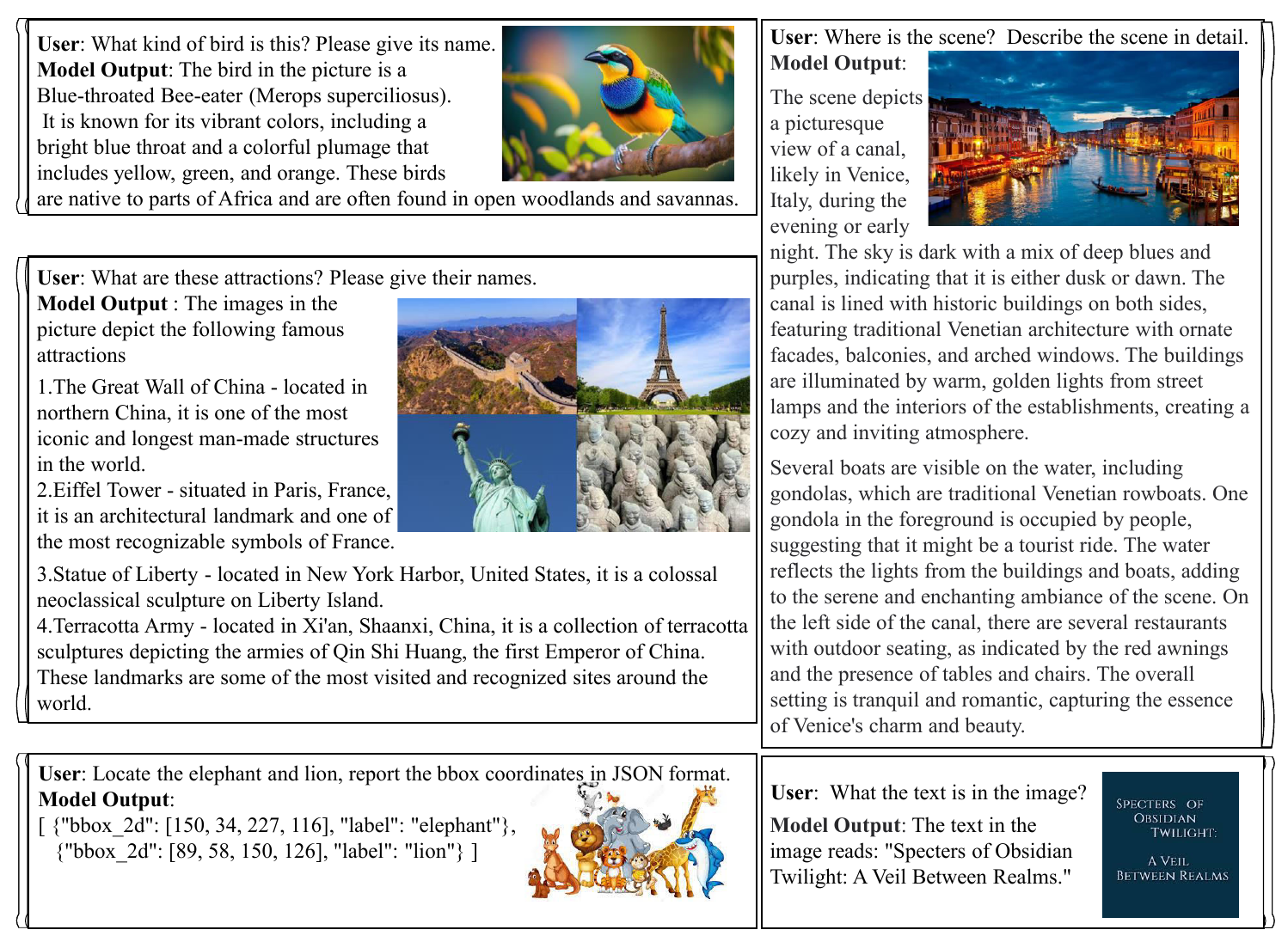}
    \caption{Qualitative examples illustrating the capabilities of UniPic2-Metaquery across diverse multimodal tasks.}
    \label{fig:udstding}
\end{figure}

\paragraph{Multimodal Understanding.}\label{subsec:mm_understand}
The Qwen2.5-VL branch remains frozen, thereby preserving its multimodal understanding capabilities. As shown in \figref{fig:udstding}, we evaluate the proposed method on six representative tasks to demonstrate its generalization ability across visual understanding domains: (a) Universal Recognition – Given an image of a bird, the model correctly classifies it as a Blue-throated Bee-eater (Merops superciliosus) and describes distinctive visual traits (e.g., bright blue throat, yellow-green-orange plumage) along with its habitat distribution. (b) Multi-instance Recognition – For a set of attraction images, the model identifies The Great Wall of China, Eiffel Tower, Statue of Liberty, and Terracotta Army, providing their geographic locations and cultural significance. (c) Scene Unserstanding– In a Venice canal scene, the model generates a detailed description capturing global layout (historic buildings flanking the canal), architectural details (ornate facades, balconies, arched windows), illumination (warm golden street lights), and dynamic elements (boats, gondolas, outdoor seating), reflecting ambient atmosphere. (d) Object Grounding with structured output – The model localizes specific cartoon animals (elephant, lion) and outputs their bounding boxes in JSON format, demonstrating the ability to produce machine-readable structured predictions. (e) OCR – The model transcribes stylized text from the image, yielding ``Specters of Obsidian Twilight: A Veil Between Realms'' with high fidelity. These results highlight the model’s capacity to handle heterogeneous input–output formats, perform fine-grained and scene-level reasoning, and maintain high accuracy in both naturalistic and stylized visual contexts.

\subsection{Ablation Studies}
\label{subsec:aba}

\begin{table}[t]
\centering
\begin{minipage}{0.48\linewidth}
\centering
\caption{Different image conditioning strategies.}
\label{tab:image_cond}
\small
\setlength{\tabcolsep}{5pt} %
\scalebox{0.83}{
\begin{tabular}{c c c c c}
\toprule
\multirow{2}{*}{\textbf{\#}}  & \multirow{2}{*}{\textbf{Model}} 
& \multicolumn{2}{c}{\textbf{Image Condition}} 
& \multirow{2}{*}{\textbf{GEdit-EN}} \\
& & \textbf{MLLM} & \textbf{DiT} &  \\
\midrule
1 & UniPic2-SD3.5M-Kontext & \xmark & \cmark & 6.31 \\
\midrule
2 & \multirow{3}{*}{UniPic2-MetaQuery} 
& \cmark & \xmark & 5.00 \\
3 &  & \xmark & \cmark & 6.40 \\
4 &  & \cmark & \cmark & \textbf{6.90} \\
\bottomrule
\end{tabular}
}
\end{minipage}
\hfill
\begin{minipage}{0.48\linewidth}
\centering
\caption{Freezing or fine-tuning (FT) the connector and DiT for UniPic2-MetaQuery.}
\label{tab:freeze_or_unlock}
\small
\setlength{\tabcolsep}{5pt}
\begin{tabular}{l c c c c}
\toprule
\# & \textbf{Connector} & \textbf{DiT} 
& \textbf{GenEval} & \textbf{GEdit-EN} \\
\midrule
1 & Freeze & FT  & 0.84 & 6.75  \\
2 & FT & Freeze  & 0.85 & 6.46 \\
3 & FT & FT  & \textbf{0.86} & \textbf{6.90} \\
\bottomrule
\end{tabular}
\end{minipage}
\end{table}

\paragraph{Pre-Training.} In Tab.~\ref{tab:image_cond}, we compare different image conditioning strategies for image editing. The result on GEdit-EN is used as the performance indicator. As illustrated in Tab.~\ref{tab:image_cond}(\#2), the image editing performance decreases drastically when reference images are only fed to the MLLM, highlighting the importance of image conditions in the diffusion process to preserve image structure and texture details. Besides, by comparing results in Tab.~\ref{tab:image_cond}(\#1 \& \#3), we observe that using MLLM as the text encoder slightly improves the performance on GEdit-EN when reference images are only passed to the DiT. Finally, we obtain the best performance by passing reference images to both the MLLM and the DiT as shown in Tab.~\ref{tab:image_cond}(\#4). In Tab.~\ref{tab:freeze_or_unlock}, we study the impact of freezing or fine-tuning the connector and DiT in the training of UniPic2-MetaQuery. As shown in Tab.~\ref{tab:freeze_or_unlock}(\#3), the best performance is achieved when unlocking both the connector and DiT.

\paragraph{Post-Training.}

We present the ablation study of different reward signals used in the post-training of UniPic2-SD3.5M-Kontext, as shown in Tab.~\ref{tab:ablation_study}. The results include the individual and combined use of GenEval reward for text-to-image reinforcement, and Skywork-EditReward and online GPT-4.1 evaluation for image editing, providing empirical insights into effective reward design for unified generation and editing models.

We use ``w/o GRPO'' as the baseline, corresponding to the UniPic2-SD3.5M-Kontext model before reinforcement learning is applied. When reward signals for image editing are introduced separately, the model’s editing performance improves significantly. Using Online GPT-4.1 Evaluation as the editing reward boosts GEditBench-EN to 6.55 and ImgEdit to 4.04, demonstrating that multi-dimensional assessment based on vision-language models (VLMs) effectively guides fine-grained editing. Furthermore, when replacing GPT-4.1 with our self-trained Skywork-EditReward model, GEditBench-EN increases to 6.59, indicating its stronger discriminative capability in evaluating editing fidelity. Notably, when only the editing task is reinforced, the T2I generation performance remains stable at 0.83 in terms of GenEval, while DPG-Bench even shows a slight improvement. This confirms the absence of negative transfer between tasks and suggests potential positive cross-task generalization.

For T2I reinforcement, introducing the GenEval reward raises the GenEval score from 0.83 to 0.87, with a modest gain on DPG-Bench, validating the effectiveness of verifiable rewards in generating images with complex semantic structures. Most importantly, when both editing and T2I tasks are reinforced, the model achieves the best overall performance: GenEval reaches 0.89, GEditBench-EN peaks at 6.59, and ImgEdit attains 4.00. This result demonstrates that our proposed PDTR strategy enables positive cross-task synergy rather than mutual interference—confirming the feasibility of joint optimization in a unified architecture. Furthermore, when comparing the two editing reward modules within PDTR, Skywork-EditReward consistently outperforms online GPT-4.1 evaluation across all benchmarks. This highlights its superior stability, higher training efficiency, and elimination of costly API calls, making it a more suitable and scalable reward module for post-training in unified multimodal systems.

In summary, the ablation study fully validates the effectiveness of our reward design and the PDTR strategy: each reward component contributes significantly to its target task, and under progressive training, the two capabilities co-evolve without conflict—providing a reliable and scalable pathway for unified image generation and editing.

\begin{table}[htbp]
\centering
\caption{Ablation Study of Reward Signals on UniPic2-SD3.5M-Kontext.}
\label{tab:ablation_study}
\small
\setlength{\tabcolsep}{5pt} % 紧凑列间距
\begin{tabular}{c c c c c c c}
\toprule
\multirow{2}{*}{\textbf{Model}} 
& \multicolumn{2}{c}{\textbf{RL}} & \multicolumn{2}{c}{\textbf{Generation}} & \multicolumn{2}{c}{\textbf{Editing}} \\
% \cmidrule(lr){2-3} \cmidrule(lr){4-5} \cmidrule(lr){6-7}
& \textbf{T2I} & \textbf{Edit} 
& \textbf{GenEval} & \textbf{DPG} 
& \textbf{GEdit-EN} & \textbf{ImgEdit} \\
\midrule
w/o GRPO & \xmark & \xmark & 0.83 & 83.68 & 6.31 & 3.95 \\
\midrule
w/ GPT-4.1 & \xmark & \cmark & 0.83 & 84.15 & 6.55 & 4.04 \\
w/ EditReward & \xmark & \cmark & 0.83 & 83.99 & 6.59 & 4.00 \\
w/ GenEval & \cmark & \xmark & 0.87 & 83.97 & 6.40 & 3.95 \\
\midrule
w/ GPT-4.1 + GenEval & \cmark & \cmark & 0.89 & 84.23 & 6.54 & 3.99 \\
w/ EditReward + GenEval & \cmark & \cmark & 0.89 & 84.36 & 6.59 & 4.00 \\
\bottomrule
\end{tabular}
\end{table}

\subsection{Failure Cases}
\label{subsec:fail}
\begin{figure}
    \centering
    \includegraphics[width=1\linewidth]{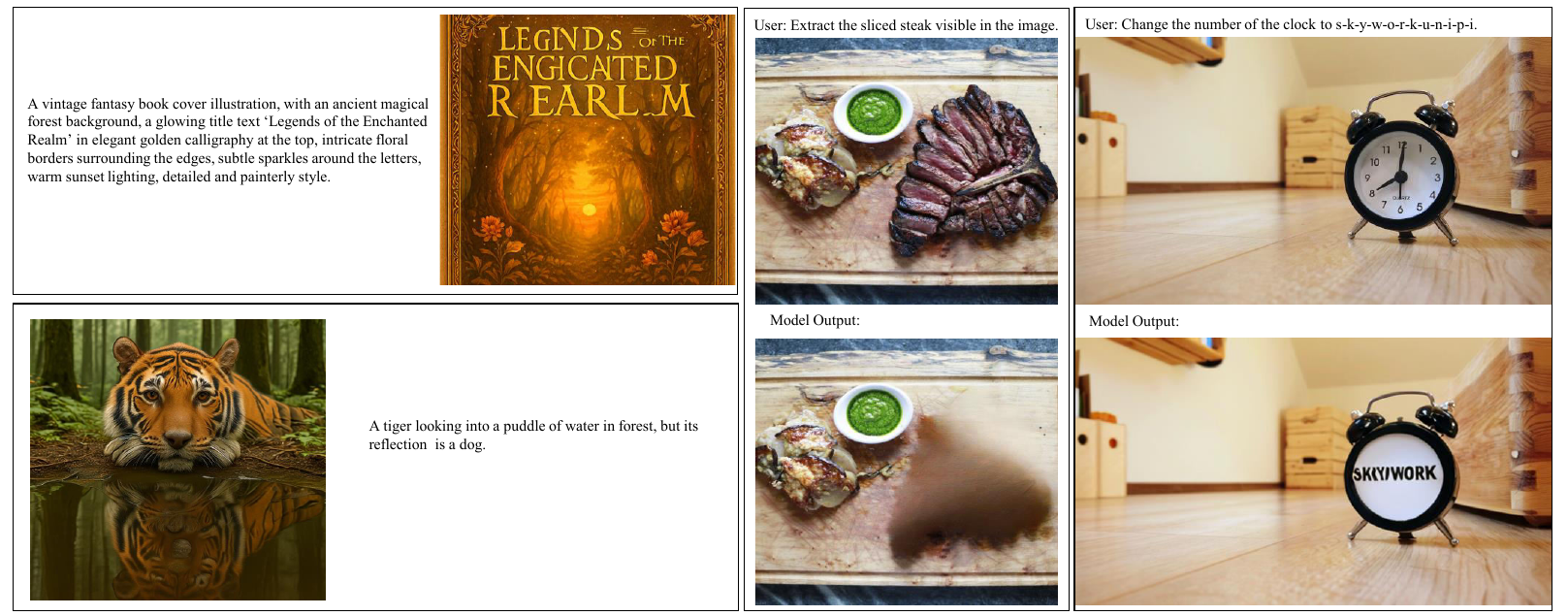}
    \caption{Failure cases on lengthy prompts with intricate semantics, complicated text and detailed object count requirements, often in long and descriptive text.}
    \label{fig:fail}
\end{figure}

\figref{fig:fail} illustrates several scenarios where our model struggles to precisely follow user instructions:
(a) Complex textual rendering – For the fantasy book cover ``Legends of the Enchanted Realm'', while the style and composition match the description, the text suffers from character substitution and distortion, a common limitation in text rendering.
(b) Special object substitution – In the ``tiger with a dog reflection'' task, although the reflection effect is plausible, semantic consistency and precision in depicting the reflected dog remain imperfect.
For image editing, operations such as object extraction, complex text modification and precise attribute conversion are also shortcomings of our model.
(a) Object extraction – When asked to extract the sliced steak, the model fails to cleanly isolate the target region, producing blurred artifacts instead of a precise cutout.
(b) Complex text modification – Changing the clock numbers to ``s-k-y-w-o-r-k-u-n-i-p-i'' yields partial success in replacing digits with text but with distortion and incompleteness issues, indicating difficulties in high-fidelity text insertion.
These cases highlight persistent challenges for contemporary text-to-image and editing models, particularly in specific count generation, complex text synthesis and simultaneous precise modifications mentioned above. As noted, the performance in such tasks could be improved through data scaling.

\section{Conclusions}
This work presents Skywork UniPic 2.0, an efficient unified multimodal understanding, generation, and editing framework built upon the SD3.5-Medium and Qwen2.5-VL architectures.
Through architectural refinements, large-scale pretraining, joint training, and the novel PDTR strategy, our approach achieves a synergistic breakthrough in both image generation and image editing.

Compared to UniPic 1.0, which was trained from scratch as an autoregressive unified model, UniPic 2.0 follows a fundamentally different paradigm: it integrates mature multimodal LLMs and pretrained diffusion generators via a parameter-efficient connector-based design. This shift enables us to preserve the strengths of the underlying models, drastically reduce training cost, and still deliver superior performance. We first develop UniPic2-SD3.5M-Kontext, a lightweight model with only 2B generation parameters, which surpasses the performance of much larger models while offering significantly faster inference. We then introduce UniPic2-Metaquery, which seamlessly unifies understanding, generation, and editing, demonstrating strong extensibility and generalization across diverse multimodal tasks. Extensive experiments show that Skywork UniPic 2.0 achieves state-of-the-art performance across benchmarks in instruction following, editing consistency, and generation stability, while maintaining high efficiency and resource friendliness. This work provides a practical, scalable, and reproducible paradigm for advancing efficient, deployable multimodal intelligence.

% \clearpage
\section{Contributors}
\textbf{Core contributors:} Hongyang Wei$^*$, Baixin Xu$^*$, Hongbo Liu$^*$, Size Wu$^{\dag}$, Jie Liu, Yi Peng, Peiyu Wang, Zexiang Liu, Jingwen He, Yang Liu$^{\ddag}$, Xuchen Song$^{\ddag}$, Yangguang Li$^{\ddag}$\\

\textbf{Contributors:} Yidan Xietian, Chuanxin Tang, Zidong Wang, Yichen Wei, Liang Hu, Boyi Jiang, Wei Li, Ying He, Yahui Zhou\\

$^*$ Equal contribution.\\
$^{\dag}$ Project Lead.\\
$^{\ddag}$ Corresponding author.

\clearpage
\bibliography{main}
\bibliographystyle{plain}
\end{document}